\pdfoutput=1
\documentclass[conference]{IEEEtran}
\IEEEoverridecommandlockouts

\usepackage{cite}
\usepackage[utf8]{inputenc}
\usepackage[T1]{fontenc}
\usepackage{amsmath,amssymb,amsfonts}
\usepackage{algorithmic}
\usepackage{graphicx}
\usepackage{textcomp}
\usepackage{xcolor}
\usepackage{booktabs}
\usepackage{array}
\usepackage{tabularx}
\usepackage{nicefrac}
\usepackage{microtype}
\usepackage{url}
\usepackage{balance}
\usepackage{hyperref}
\hypersetup{
  hidelinks,
  breaklinks=true,
  pdftitle={From Output Quality to Auditability: A Data Mining Agenda for LLM-Generated Artifacts},
  pdfauthor={Mingjia Qian and Limeng Ge}
}
\providecommand{\citep}[1]{\cite{#1}}
\providecommand{\citet}[1]{\cite{#1}}

\begin{document}

\title{From Output Quality to Auditability:\\ A Data Mining Agenda for LLM-Generated Artifacts\thanks{\copyright~2026 IEEE. Personal use of this material is permitted. Permission from IEEE must be obtained for all other uses, in any current or future media, including reprinting/republishing this material for advertising or promotional purposes, creating new collective works, for resale or redistribution to servers or lists, or reuse of any copyrighted component of this work in other works.}}

\author{\IEEEauthorblockN{Mingjia Qian}
\IEEEauthorblockA{\textit{Department of Sociology}\\
\textit{Peking University}\\
Beijing, China\\
\texttt{qianmingjia26@stu.pku.edu.cn}}
\and
\IEEEauthorblockN{Limeng Ge}
\IEEEauthorblockA{\textit{Physical Sciences Division}\\
\textit{University of Chicago}\\
Chicago, United States of America\\
\texttt{limengge@uchicago.edu}}
}

\maketitle

\begin{abstract}
Large language models increasingly emit complex artifacts that include claims, code, citations, experiments, and tool traces, and these artifacts enter scientific and organizational workflows faster than they can be checked. Such artifacts create a data mining problem: their audit structure can be shaped by the system under review to appear trustworthy and must be inspected under bounded human oversight. This paper develops a data-mining agenda for \emph{AI-generated audit artifacts} rather than a complete audit system. It treats these artifacts as data objects represented by verification graphs and frames auditability through \emph{verification cost}: the representation, checking, and trust required for audit. We define four benchmarkable tasks: audit-graph extraction, budget-constrained verifier routing, adversarial-auditability detection, and audit-trace and cost-drift mining, each specified by inputs, outputs, metrics, candidate data, and baselines. The goal is to make verification cost an explicit, mineable, and contestable quantity for generated artifacts.
\end{abstract}

\begin{IEEEkeywords}
AI-generated artifacts, auditability, data mining, large language models, verification cost
\end{IEEEkeywords}

\section{Introduction}
Data mining has long included synthetic and simulated data as objects of analysis. In the LLM era, the main change is not the existence of generated data but the role of generated artifacts: they now carry claims, code, citations, and experimental traces into analytic, scientific, and organizational workflows. Once these artifacts enter downstream use, output quality alone is insufficient. Their claims must also be represented in a form that can be checked efficiently under limited human oversight. This paper treats \emph{audit-ready data artifacts} as structured data objects and \emph{verification cost} as a data mining objective. An audit-ready artifact links claims to evidence and provenance, records relevant process traces and verification operations, and makes residual risks explicit for selective audit under a fixed budget. We distinguish three components of verification cost: \emph{representation cost}, the effort required to make an artifact checkable; \emph{checking cost}, the resources needed to perform the relevant checks; and \emph{trust cost}, the effort required to assess the reliability and coverage of those checks. The resulting task is to allocate heterogeneous verification resources across claims and evidence within explicit layer-specific budgets.

Current LLM systems increasingly operate as research and engineering agents. AI Scientist and Agent Laboratory illustrate automated research workflows \citep{lu2024aiScientist,schmidgall2025agentLaboratory}, while MLE-bench and AstaBench evaluate agents that chain code, experiments, reports, and tool traces into pipelines \citep{shern2024mleBench,bragg2026astabench}. As these artifacts enter analysis, review, and governance, their verifiable structure and verification cost become mining targets. A central risk is proxy-induced audit failure: even cooperative systems may emit artifacts that look auditable without becoming more truthful.

The paper makes three contributions. First, it identifies \emph{AI-generated audit artifacts} as endogenous data objects whose audit structure may be shaped by the system under review to satisfy extractors, verifiers, benchmarks, or reviewers. Second, it represents them as verification graphs and decomposes verification cost into representation, checking, and trust costs, with formal verifiers, executable checks, and LLM judges treated as verification operators. Third, it defines four benchmarkable mining tasks: audit-graph extraction, budget-constrained verifier routing, adversarial-auditability detection, and audit-trace and cost-drift mining.

\section{Why This Is a Data Mining Problem}

LLM-generated artifacts are becoming complex, semi-structured data. A single agent transcript can contain natural-language claims, source code, tool calls, test outputs, and multi-step reasoning traces, among other elements. These are heterogeneous, linkable, partially structured objects: the natural terrain of data mining. The opportunity is to mine \emph{verifiable structure} from such artifacts so that bounded humans and automated verifiers can inspect the right claims, evidence, and dependencies without reconstructing the whole artifact from prose.

The connection to existing data mining research is direct but not mechanical. Information extraction, provenance, factuality evaluation, scalable oversight, and verifier systems provide useful components, but do not jointly represent artifact-level support, verification cost, and residual risk. Yu and Jia study resource-bounded projections of machine-scale mathematical proofs, where formal validity supplies a comparatively fixed target \citep{yu2025truth}; Nimase et al. study an explanation--extraction conflict between service providers and users \citep{nimase2025navigating}. Our premise is instead endogeneity, with verification conditioned on an explicit audit specification. The agenda places heterogeneous claims, evidence, provenance, verifier routing, and cost drift within a common data object.

The key difference is endogenous audit structure. Records, logs, citations, or annotations may be noisy in conventional settings, but they are not usually optimized for the audit protocol itself. Here, claims, citations, traces, rationales, and verifier-facing metadata are produced by or derived from the system under review and may be shaped to satisfy extractors, benchmarks, verifiers, or reviewers. The task is therefore to mine verifiable structure under bounded, heterogeneous oversight, measuring success by representation, checking, and trust costs, as well as coverage and residual risk.

When generation is cheap but rigorous checking is expensive, evaluation creates an information asymmetry between model outputs and human evaluators. As in Akerlof's market for lemons, hard-to-inspect quality can be displaced by cheap signals of quality \citep{akerlof1970market}. Fluent, authoritative, or citation-shaped answers may then be rewarded over better-grounded but harder-to-verify ones. As benchmarks saturate, marginal gains may likewise reflect shallow proxies rather than lower verification cost \citep{goodhart1984problems,agrawal2024references,factor2024}. This accumulates cognitive technical debt: systems generate dense artifacts faster than humans can audit them \citep{sculley2015hidden}, potentially eroding the human data pipelines that oversight depends on \citep{santy2025incentives}. Treating verification cost as a measurable target provides one response.

\section{Audit-Ready Data Artifacts}
\label{sec:vea}

We now make verification cost operational through audit-ready artifacts, verification graphs, and cost metadata.

\subsection{A Running Example}

Consider an illustrative automated-research artifact: an agent reports \textbf{58.1} on the MATH test split, up from a \textbf{52.3} baseline, a \textbf{+5.8} improvement. The scores, however, use different data: \texttt{test} for the method and a stale or mismatched \texttt{test\_v1} for the baseline \citep{neurips2024-ying-automatingdatasetupdates}. The load-bearing flaw is the comparison, not either number. Free-form prose may lead a reviewer to rerun the method yet miss the split mismatch. An audit-ready version instead decomposes the claim, localizes numbers and configurations, records their splits, and exposes residual risks. A policy can then route metric-consistency and split-alignment checks before paying for reproduction (Table~\ref{tab:example}).

\begin{table}[!t]
\centering
\small
\caption{A benchmark-improvement claim represented as free-form prose versus as a concrete audit-ready instance. Numbers are illustrative.}
\label{tab:example}
\renewcommand{\arraystretch}{1.15}
\setlength{\tabcolsep}{3pt}
\renewcommand{\tabularxcolumn}[1]{m{#1}}
\begin{tabularx}{\linewidth}{>{\raggedright\arraybackslash}m{0.28\linewidth} >{\raggedright\arraybackslash}X >{\raggedright\arraybackslash}X}
\toprule
\textbf{Dimension} & \textbf{Free-form report} & \textbf{Audit-ready artifact} \\
\midrule
Claim structure & ``Our method improves benchmark score.'' & $c_1$: method scores 58.1 on \texttt{test}; $c_2$: baseline scores 52.3 on its split; $c_3$: $c_1-c_2$ is a $+5.8$ gain, depending on $c_1,c_2$. \\
Support structure & Numbers, configs, and logs are scattered across prose and appendix. & $e_1$: metric line 0.581; $e_2$: baseline line 0.523; $e_3$: config split=\texttt{test}; $e_4$: baseline split=\texttt{test\_v1}; $e_5$: run and commit IDs. \\
Verification route & Reviewer reruns the method and reconfirms 58.1. & Metric check on $e_1,e_2$; split-alignment check on $e_3$ vs.\ $e_4$; optional reproduction from $e_5$. \\
Residual risk & Risks are implicit in the narrative. & Fires on $c_3$: invalid baseline / split leakage, detected as $e_3\neq e_4$. \\
\bottomrule
\end{tabularx}
\end{table}

\textbf{Audit trace.} Constructing $A=(C,E,L,D,P,M)$ is \emph{audit-graph extraction} (Table~\ref{tab:agenda}). Suppose checking budget $B_C$ permits opening five evidence units or one reproduction. A cheap-first policy $\pi$ first confirms that $e_1,e_2$ match 58.1 and 52.3, then compares $e_3,e_4$ and finds \texttt{test} versus \texttt{test\_v1}. In this illustration, the invalid-baseline risk on $c_3$ is flagged in about three seconds, before a roughly 40 GPU-minute reproduction. The free-form path instead reconfirms 58.1 and misses the flaw, giving $\text{AuditSuccess@}B\!:\,0\to 1$ and the cost profile $(M_R,M_C,M_T)$.

\subsection{The Verification Graph}

Given an LLM-generated artifact $Y$ and an audit specification $S$ that fixes the target property, admissible evidence, temporal and domain context, verifier pool, and tolerated residual risk, the goal is to construct an audit artifact
\[
A=(C,E,L,D,P,M),
\]
where $A$ is a data mining object rather than a formatting wrapper around text. Its components are:
\begin{itemize}
    \item $C$, a set of \emph{atomic claims} with residual-risk attributes: minimal self-contained statements represented for local checking, which may remain mutually dependent through $D$ \citep{min2023factscore}.
\item $E$, a set of \emph{localized evidence units} such as citations, retrieved passages, code outputs, test results, and log lines, with source, timestamp, and context metadata where applicable \citep{gao2023enabling}.

\item $L \subseteq C \times E \times \mathcal{T}_L$, where $\mathcal{T}_L$ contains evidential-link types for support, contradiction, contextualization, and supersession \citep{reclaim2025,laquer2025}.

\item $D \subseteq C \times C \times \mathcal{T}_D$, where $\mathcal{T}_D$ contains claim-dependency types for prerequisites, comparisons, and aggregations.

\item $P$, \emph{provenance and process traces} such as tool calls, data transformations, seeds, configurations, and execution logs.

\item $M=(M_R,M_C,M_T)$, \emph{verification-cost metadata}: representation-cost metadata, checking-cost metadata, and trust-cost metadata, including pre-routing estimates and post-routing verifier summaries.
\end{itemize}
Since bounded oversight prevents an auditor from inspecting every node or rerunning every step, the graph supports verifier routing and selective audit. Verification is relative to $S$, rather than to a universal standard of real-world truth; for subjective or normative claims, the graph may expose provenance and procedural compliance without certifying substantive correctness.
Operationally, $S$ should be declared before auditing rather than inferred after a result is seen. Different stakeholders may instantiate different specifications--for example, a scientist may prioritize reproducibility, whereas a regulator may prioritize provenance and temporal validity. A benchmark should therefore publish $S$ with each task and report sensitivity to plausible alternative specifications. This makes disagreement inspectable: two auditors can differ because they used different evidence rules or risk tolerances, rather than because ``verified'' silently changed meaning.
The mining task is to construct $A$ from $Y$ and accessible sources such as evidence corpora, logs, repositories, and test outputs. Because the three cost layers need not share units, the objective is budget-constrained audit success rather than an unqualified scalar sum:
\[
\begin{aligned}
\max_{\pi}\quad &\mathbb{E}[\mathrm{AuditSuccess}(A,Y,\pi;S)]\\
\text{s.t.}\quad &\mathrm{Cost}_{j}(A,Y,\pi)\leq B_j,\quad j\in\{R,C,T\},\\
&\mathrm{Faith}(A,Y;S)\geq\tau,
\end{aligned}
\]
where $B=(B_R,B_C,B_T)$ is a vector budget, and AuditSuccess records correct disposition within that budget by detecting a labeled load-bearing flaw or accepting a clean artifact, while false-positive rate is reported separately. Here $\pi$ routes verifiers to claims, while $\mathrm{Faith}$ measures whether claims, typed links, dependencies, traces, and verifier reports preserve the artifact's support structure. Scalarization is appropriate only when the audit protocol preregisters commensurate units or conversion weights.

The graph is deliberately not a guarantee of truth. Extracted claims can drop context, evidence links can be wrong, verifier outputs can under-cover the relevant behavior, and generated rationales need not reflect the process that produced the answer \citep{turpin2023language,wei2022chain}. This is why \emph{graph faithfulness} and verifier trust are measured rather than assumed. In practice, graphs can be built by post-processing free-form outputs, co-generating text and graph together, or instrumenting agent workflows so that evidence, verifier invocations, and process traces are emitted during generation.

\subsection{Verification-Cost Metadata}

Benchmarks today typically report only accuracy or a final score. But two equally accurate outputs differ if one can be represented, checked, and trusted quickly while the other requires reconstructing claims, rerunning experiments, and calibrating a weak judge. A \emph{verification-cost metadata} layer makes this cost structure observable alongside task scores (Table~\ref{tab:metadata}).

\begin{table}[!t]
\centering
\small
\caption{A verification-cost metadata layer reported alongside task accuracy. The three cost types ask whether an artifact is checkable, how expensive the check is, and why the check should be trusted.}
\label{tab:metadata}
\renewcommand{\arraystretch}{1.15}
\setlength{\tabcolsep}{3pt}
\renewcommand{\tabularxcolumn}[1]{m{#1}}
\begin{tabularx}{\linewidth}{>{\raggedright\arraybackslash}m{0.24\linewidth} >{\raggedright\arraybackslash}m{0.38\linewidth} >{\raggedright\arraybackslash}X}
\toprule
\textbf{Cost type} & \textbf{Metadata examples} & \textbf{Question} \\
\midrule
Representation cost & Extraction or instrumentation time; evidence locality; storage and update cost & Is the artifact checkable? \\
Checking cost & Human time; test runtime; reproduction compute; judge or API cost & How expensive is the check? \\
Trust cost & Labeled calibration effort; coverage analysis; specification-validation time & What does confidence in the check cost? \\
\bottomrule
\end{tabularx}
\end{table}

An existing multi-step benchmark such as MATH \citep{hendrycks2021math} can keep its accuracy axis and add a cost profile $(Acc,M_R,M_C,M_T)$, where each $M_\cdot$ is a vector of task-specific metadata. The extraction, instrumentation, storage, and maintenance needed to sustain this layer are counted as representation cost rather than treated as free infrastructure. Graph faithfulness, verifier coverage, calibration error, and residual risk are reported as quality or risk outcomes rather than cost units. Systems can then be compared by budgeted audit success and Pareto frontiers without collapsing auditability into a single scalar. Existing process-supervision and generative-verifier results suggest such signals can become optimization targets, not just reports \citep{iclr2024-lightman-letsverify,iclr2025-zhang-generativeverifiers}.

\section{Research Directions, Challenges, and Success Criteria}

Table~\ref{tab:agenda} turns the agenda into concrete mining tasks with inputs, outputs, metrics, candidate data, and baselines.

\begin{table*}[!t]
\centering
\footnotesize
\caption{Concrete mining tasks for audit-ready artifacts.}
\label{tab:agenda}
\renewcommand{\arraystretch}{1.12}
\setlength{\tabcolsep}{2pt}
\begin{tabularx}{\textwidth}{>{\raggedright\arraybackslash}m{0.18\textwidth} >{\raggedright\arraybackslash}m{0.18\textwidth} >{\raggedright\arraybackslash}m{0.17\textwidth} >{\raggedright\arraybackslash}m{0.17\textwidth} >{\raggedright\arraybackslash}X}
\toprule
\textbf{Task} & \textbf{Input} & \textbf{Output} & \textbf{Metrics} & \textbf{Candidate data / baselines} \\
\midrule
Audit-graph extraction & Generated artifact $Y$; evidence, logs, code, and traces & Verification graph $A=(C,E,L,D,P,M)$ & Claim F1; typed-link P/R; critical-dependency recall; graph faithfulness; representation cost & RAGTruth, ALCE, FActScore, MLE-bench logs; prompt extraction, NLI alignment \\
Budget-constrained verifier routing & Verification graph; verifier pool; budget $B$ & Policy $\pi$ routing claims to tests, proof checkers, reproduction, or judges & AuditSuccess@$B$; false-positive rate; time to verification; cost--coverage curve & MATH/PRM-style data, code benchmarks, MLE-bench; verify-all, random, uncertainty routing \\
Adversarial-auditability detection & Artifact with accidental, optimization-induced, or deliberate audit failures & Failure type, flagged links or traces, calibrated risk & Macro-F1/AUC by failure type; calibration error; robustness gap; residual risk & Natural omissions; injected false support, leakage, or trace defects; NLI, LLM judge, graph anomaly detection \\
Audit-trace and cost-drift mining & Audit logs, benchmark histories, verifier outputs, review traces & Cost prediction, drift detection, verifier-overfitting signals & Cost-prediction error; drift lead time; missed-flaw rate; coverage drift & Agent traces, benchmark histories, review logs; time-series anomaly and changepoint detection \\
\bottomrule
\end{tabularx}
\end{table*}

The tasks are separable but compositional: extracted graphs feed routing, failures inform auditability detection, and accumulated traces support cost-drift mining. A dataset can pair artifacts with labeled audit failures, a vector budget, and costed verifiers \citep{microsoft2026verifier}. A practical first benchmark should combine graph extraction and routing on versioned reports whose claims, splits, configurations, evidence, and load-bearing defects admit reproducible labels before deliberate-gaming settings are introduced. Labels should identify the affected claim, dependency, failure type, and cheapest successful verifier. Evaluation can then separate extraction from routing error and compare audit success at fixed layer-specific budgets.

\textbf{From tasks to cost layers.} Audit-graph extraction targets representation cost \citep{ragtruth2024}, verifier routing targets checking cost \citep{lean2024theoremProving}, and adversarial-auditability detection targets trust cost \citep{liu2025rethinkingAutoformalization}. Audit-trace and cost-drift mining spans all three by tracking missed flaws, verifier behavior, and evolving rubrics across model versions and review protocols \citep{jetts2025}.

\textbf{A stress-test case: adversarial auditability.} The framework does not assume that generators and auditors have conflicting interests. Its premise is endogeneity: the system produces the audit-facing structure used to evaluate it. Failures may be accidental, such as incomplete evidence or verifier miscalibration; optimization-induced, such as proxy gaming or verifier overfitting; or deliberate, such as fabricated support or laundered traces \citep{drtulu2025}. Only the last category presumes adversarial intent. Unlike standard adversarial robustness, the attack surface is the generated audit structure and the failure criterion is false acceptance under a bounded protocol, rather than sensitivity to a small input perturbation \citep{bagoftricks2024,sharma2023understanding,xu2024knowledgeconflicts}. Graph faithfulness, verifier coverage, and auditability-failure detection therefore require methods that model and stress-test generated structures and the verifiers used to inspect them, rather than only final answers. Stress tests should also distinguish naturally observed failures from injected ones, so performance on conspicuous synthetic corruptions is not mistaken for robustness to ordinary omissions.

\textbf{Relation to downstream systems.} These artifacts and their metadata are a substrate that downstream training and oversight methods can consume, not a replacement for those methods. Reward models could score verification cost alongside correctness, encouraging systems to expose checkable claims and localized evidence \citep{leike2018scalable,ouyang2022training}. Prover--verifier setups could reward generators for producing artifacts whose load-bearing claims admit targeted checks \citep{irving2018ai}. Formal proof assistants illustrate how formalization can lower the checking cost of selected mathematical or program properties \citep{lean2024theoremProving}. Scalable-oversight pipelines could use verification graphs to allocate human and automated review across claims and evidence \citep{burns2023weak,kenton2024scalable}. Automated weak-to-strong research workflows provide a complementary setting for mining audit traces over time \citep{anthropic2026aar}. Across these settings, the common requirement is a data object and measurement layer exposing representation, checking, and trust costs, verifier coverage, and residual risk.

\textbf{Bounded oversight as a protocol.} Verification cost is meaningful only under a standardized protocol. The protocol should specify the auditor, verifier mix, and budget; compare outputs with and without verification graphs; include planted flaws, natural cases, and clean controls; and record representation, checking, and trust costs. Evaluation should report the claims inspected, evidence localization, flaw detection within budget, and false-positive rate. These measures, reported alongside accuracy, allow auditability to be compared across systems.

\textbf{Expected outcomes.} ``State of the art'' would no longer mean only a higher score, but a comparable score accompanied by lower representation, checking, and trust costs under stress-tested audit protocols \citep{amodei2016concrete,bowman2022measuring}.

\textbf{Limitations.} None of this removes the human from the loop, and the approach can fail in instructive ways. Gamed metadata moves the proxy problem up one level; weak extractors produce misleading graphs; misaligned formalization certifies wrong statements; under-specified tests pass unsupported claims; and poorly calibrated judges reward persuasive falsehoods. The agenda focuses on English-language artifacts. Multilingual settings may alter claim segmentation, evidence authority, entailment judgments, and verifier calibration, requiring language-specific specifications. Verification cost should therefore remain a reported, contestable profile rather than a scalar objective.

\section{Conclusion}

This framing points toward a new infrastructure layer for generated knowledge. Benchmarks could report verification costs alongside accuracy, while review workflows support representation, checking, and trust. Repositories could likewise store verification graphs, verifier outputs, and audit logs alongside generated artifacts, allowing future users to examine how a result was represented, checked, trusted, challenged, and revised.
Audit-ready artifacts neither guarantee truth nor eliminate the need for human oversight. Their value lies in lowering representation and checking costs, making trust costs and residual risks explicit, and improving the likelihood that important flaws are detected within a bounded budget. Over time, an AI-generated report without verification-cost metadata may come to resemble a dataset without documentation or a model result without an evaluation protocol. Treating verification cost as a measurable, mineable, and contestable objective can therefore make generated artifacts easier to audit without conflating auditability with truth.

\balance

\end{document}